\documentclass{article} 
\usepackage{spconf,amsmath,epsfig}
\usepackage{hyperref}
\usepackage{url}
\usepackage{graphicx}
\usepackage{multirow}
\usepackage{textcomp}
\usepackage{float} 
\usepackage{adjustbox} 

\title{Polar-VQA: Visual Question Answering on Remote Sensed Ice sheet Imagery from Polar Region}

\twoauthors
 {Argho Sarkar}
	{University of Maryland Baltimore County\\
	Department of Information Systems\\
	USA}
 {Maryam Rahnemoonfar}
	{Lehigh University\\
	Department of Computer Science and Engineering, and\\ Civil and Environmental Engineering\\
	USA}

\begin{document}

\maketitle

\begin{abstract}
\par For glaciologists, studying ice sheets from the polar regions is critical. With the advancement of deep learning techniques, we can now extract high-level information from the ice sheet data (e.g., estimating the ice layer thickness, predicting the ice accumulation for upcoming years, etc.). However, a vision-based conversational deep learning approach has not been explored yet, where scientists can get information by asking questions about images. In this paper, we have introduced the task of Visual Question Answering (VQA) on remote-sensed ice sheet imagery. To study, we have presented a unique VQA dataset, \textit{Polar-VQA}, in this study. All the images in this dataset were collected using four types of airborne radars. The main objective of this research is to highlight the importance of VQA in the context of ice sheet research and conduct a baseline study of existing VQA approaches on \textit{Polar-VQA} dataset.
\end{abstract}

\section{Introduction}
\label{introduction}
\par
\begin{figure*}[!htp]
\centering
        \includegraphics[width=.58\textwidth]{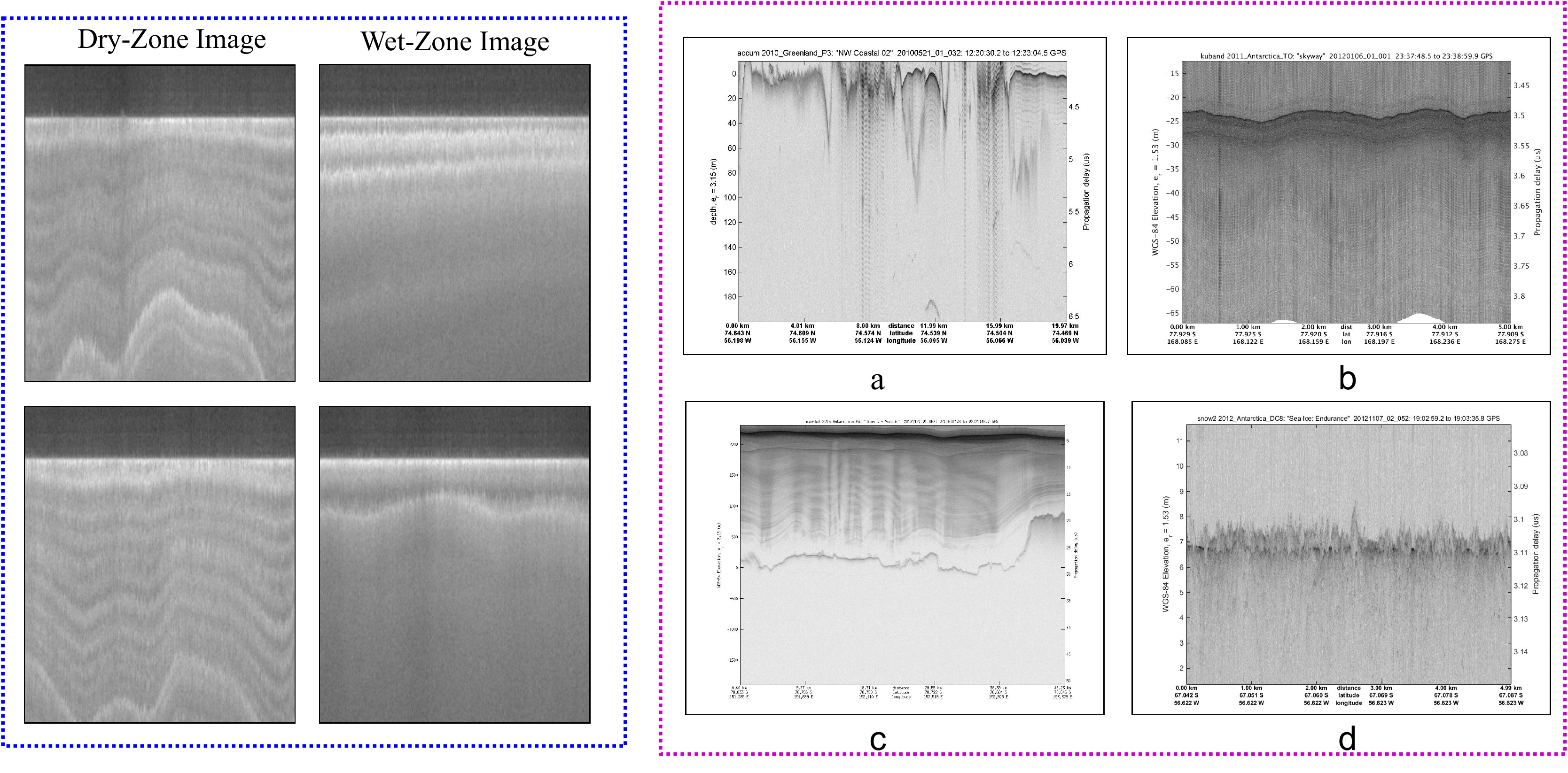}
    \caption{(1) Left figure shows the examples of images from dry and wet zone respectively, (2) right figure shows the images collected using (a) Accumulation Radar, (b) Ku Band Radar, (c) Radar Depth Sounder (d) Snow Radar}
    \ \label{Fig:radar}
\end{figure*}

 Exploring polar regions and retrieving information from the ice sheets have great significance in glaciology studies. With this aim, a large amount of data has been collected from various airborne radars, such as \textit{Accumulation Radar}, \textit{Ku-Band Radar}, \textit{Radar Depth Sounder}, \textit{Snow Radar}, over several years. All the data are primarily gathered from various geo-locations (e.g., \textit{Antarctica, Greenland}) and zones such as the \textit{dry and wet} of those locations. By definition, dry-zone imagery is collected from polar regions at a distance from the coastal area, whereas wet-zone imagery is collected from adjacent places in the coastal area. Data from these various sources and different locations can be utilized to study and monitor the ice sheets. With the advancement of many deep learning techniques, many challenges have been efficiently addressed in recent times regarding polar ice sheet research. Those researches mainly focused on estimating ice-layer thickness \cite{9378070}, ice-layer tracking \cite{varshney2021deep}, physics-driven Deep learning simulation for generating ice-imagery \cite{yari2021airborne}, etc. However, more different types of information can be extracted from the ice sheet data that are very useful for scientists to have a better understanding. Identifying the radar type, the geo-locations, and recognizing the zone from images will assist the scientist from a different point of view. For instance, while estimating the ice accumulation over years, it is very important to understand the impact of geo-location and the zone (dry or wet) on that estimation. Additionally, dry-zone images naturally have more layers than wet-zone images. Fig. \ref{Fig:radar} represents both dry and wet zone images. Scientists often find it hard to visually differentiate between dry and wet zone images. Thus, they need an intelligent opinion that can extract high-level features and provide correct information. These are the reasons why it is necessary to identify locations and zones from images. Though all the information could be stored in the metadata (i.e., where all the information regarding data collection parameters is kept), it is very time-consuming to extract information from the metadata. 

\begin{figure}[h]
\centering
        \includegraphics[scale=.5]{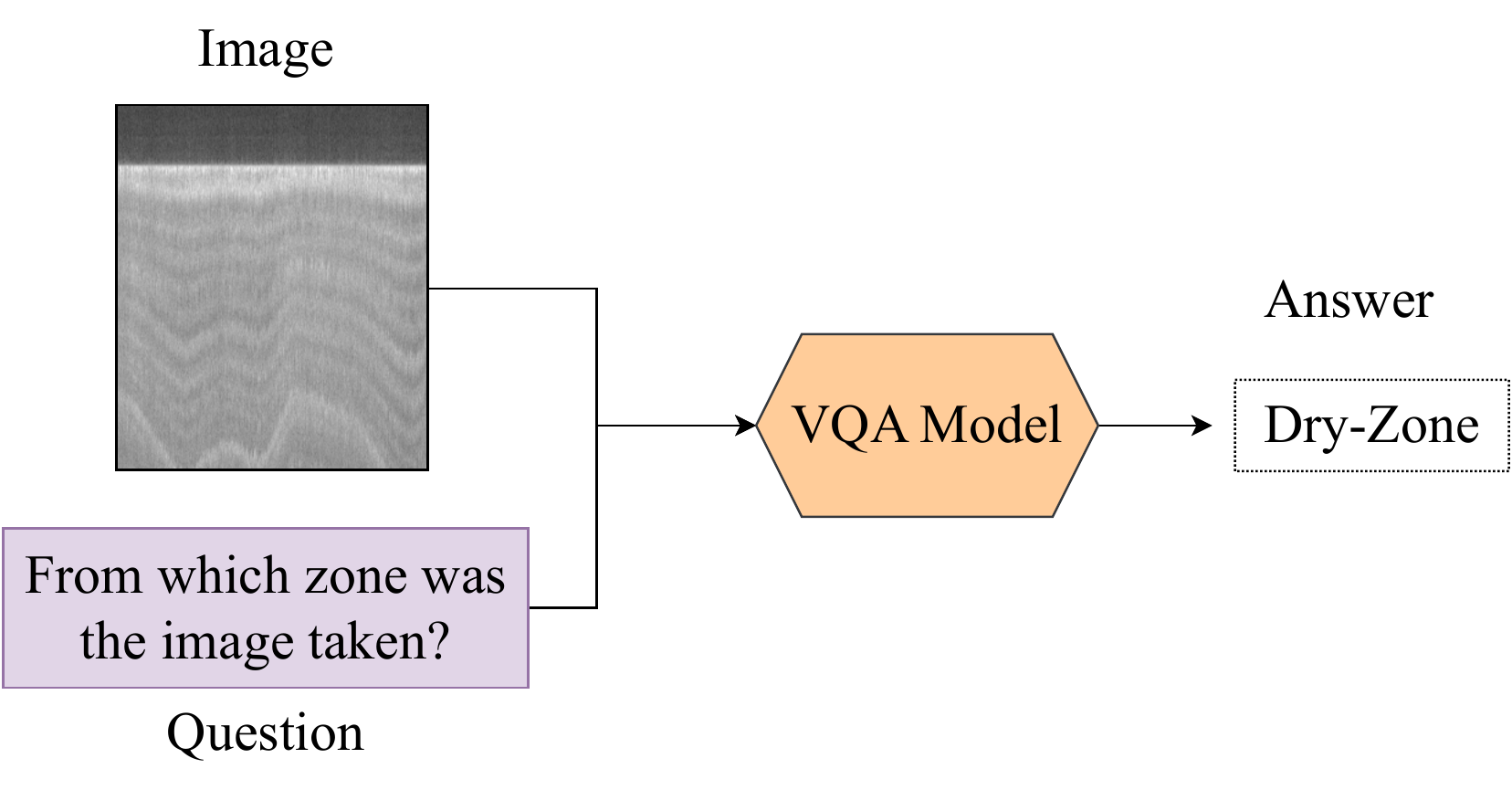}
    \caption{Visual Question Answering (VQA) framework for remote-sensed ice imagery. In this approach, an image and question are fed into the VQA model, and finally we get the answer to that question from the VQA model}
    \label{Fig:system}
\end{figure}

\par Thus, an interactive approach is needed where a scientist can get this information by asking questions about the image. ``Is the image taken from accumulation radar or snow radar? ", ``What kind of zone is this?" are some examples of questions that could be asked by an expert to get that information. To address this issue, we have presented a visual question answering framework based on ice sheet data. Fig. \ref{Fig:system} represents the VQA framework.

\par Visual question answering (VQA) is an interactive approach where we can ask questions based on images and get query-based rational responses. Many researches have been conducted \cite{antol2015vqa, yang2016stacked, yu2017multi} on several applications of the VQA approach. Most of the research is limited to natural \cite{antol2015vqa}, medical \cite{abacha2019vqa}, remote-sensing \cite{sarkar2021vqa,9897381} images. In this work, we have presented a unique VQA dataset, \textit{Polar-VQA}, on ice sheet data for the first time. The main focus while developing this dataset is to identify three main components from the images: locations, radar-type and zone through asking questions. These details are not visible within the collected images and are also difficult for a human to visually differentiate. As VQA is able to provide high-level scene understanding, we take advantage of this deep learning technique in our ice sheet research. In addition, there is no other deep-learning techniques that can provide these three types of information per image simultaneously. Image classification, segmentation \cite{doshi2018satellite}, and object detection \cite{chen2018benchmark}  are hardly suitable for this application. To the best of our knowledge, this is the first VQA application for ice sheet research.

\section{Polar-VQA Dataset}
\label{dataset}
\subsection{Data Collection Process}
\par All ice sheet data are collected in image format using four different types of radars: accumulation radar, ku band radar, radar depth sounder, and snow radar. Fig. \ref{Fig:radar} shows the outcome of four type of these radar images. These airborne sensors are installed in the Uninhabited Aircraft System (UAS), which provides an aerial platform for ice-penetration. All the data are available at the Center for Remote Sensing of Ice Sheets (CReSIS) \url{https://data.cresis.ku.edu/data}. For our experiment, we only considered the data collected from Antarctica and Greenland. 

\subsection{Type of Questions}

\par All the questions in our dataset mainly focus on identifying the type of sensors, the geo-locations of the ice images, and recognizing the dry and wet zone images, respectively. Thus, all the questions are classified into three categories: sensor category, place category, and zone category. In addition, all these categories are subdivided into two types of questions

\begin{itemize}
    \item  \textbf{Open-Ended (OE)}: Questions in this category start with the ``WH" word. It allows the VQA system to give free-form responses. ``What type of sensor is this?", ``From which zone was the image taken?" are examples of this question category.
    \item \textbf{Closed-Ended (CE)}: Close-ended questions usually offer a VQA system with a limited number of answers, such as `yes/no' or any multiple-choice answer. ``Is this data from Antarctica or Greenland?", ``Is the image taken from snow radar?" are some of the examples of this kind.
\end{itemize}

\section{Baseline Studies}
\label{exp}
In this study, we conduct baseline comparative analysis by considering the five baseline VQA models on \textit{Polar-VQA} dataset. 

\begin{itemize}

    \item In \textit{Question-only} baseline model, the main purpose is to investigate the performance of predicting answer directly from questions without seeing any image. 
    \item \textit{CNN+LSTM} \cite{Lu2015} is a coarse VQA approach  where image features, extracted by CNN, and question features, are extracted from LSTM, are combined by element-wise multiplication and fed into the classifier to predict the answer.
    \item \textit{VIS+LSTM} \cite{ren2015image} is a simple VQA approach . Firstly, a image is fed into a convolution layers (CNN) and take the output. Then this output is considered for initializing the LSTM layers which then try to predict the answers. 
    \item \textit{Stacked Attention Network (SAN)} \cite{yang2016stacked} is a attention-based VQA model. In this work-frame, multi-step visual attention is considered to predict the answer.
    \item \textit{Multi-modal Factorized Bilinear Pooling with Co-Attention (MFB+CoAtt)} \cite{yu2017multi} is another kind of attention-based VQA model.  Considering attention for both image and question level is the main characteristic of this model. In the fusion stage, image and question features are combined  with multi-modal factorized bilinear pooling block described in \cite{yu2017multi}.
\end{itemize}

\begin{table*}[!htp]
\centering
\caption{ Accuracy Comparison between Baseline VQA Models on test data}
\vspace{.5cm}
\begin{tabular}{ccccc} 
VQA Model & Overall & Radar Category & Place Category & Zone Category \\ \hline \hline
Q-Only        & 0.51 & 0.63 & 0.54 & 0.43 \\ 
VIS-LSTM \cite{ren2015image}       & 0.96 & 0.96 & 0.93 & 0.99 \\
CNN+LSTM \cite{Lu2015}      & 0.96 & 0.95 & 0.94 & 0.99 \\
SAN \cite{yang2016stacked} & \textbf{0.97}                        & \textbf{0.97}           & \textbf{0.94}           & \textbf{0.99}          \\
MFB-CoAttention \cite{yu2017multi} & 0.97 & 0.96 & 0.94 & 0.99\\ \hline
\end{tabular}
\label{tab:accu}
\end{table*}

\begin{table}[h]
\centering
\caption{Question-wise Accuracy Comparison Between Baseline Models for Each Question Category}
\begin{tabular}{lcccc}
                       &        VQA Model        & \multicolumn{1}{l}{Open-Ended} & Close-Ended \\ \hline \hline
\multirow{4}{*}{Zone}  & VIS-LSTM        & 1                                       & 0.995                \\
                       & CNN+LSTM        & 1                                       & 0.998                \\
                       & SAN             & 1                                       & 0.995                \\
                       & MFB-CoAttention & 1                                       & 0.996                \\ \hline
\multirow{4}{*}{Place} & VIS-LSTM        & 0.93                                    & 0.93                 \\
                       & CNN+LSTM        & 0.94                                    & 0.94                 \\
                       & SAN             & 0.93                                    & 0.94                 \\
                       & MFB-CoAttention & 0.93                                    & 0.94                 \\ \hline
\multirow{4}{*}{Radar} & VIS-LSTM        & 0.99                                    & 0.96                 \\
                       & CNN+LSTM        & 0.99                                    & 0.94                 \\
                       & SAN             & 0.99                                    & 0.96                 \\
                       & MFB-CoAttention & 0.99                                    & 0.95            \\ \hline    
\end{tabular}
\label{tab:accu_2}
\end{table}

\par  We have considered the batch size 16, \textit{adam} optimizer is taken as a optimizer and \textit{early stopping} criteria with patience 10 is considered for all the experiments. \textit{ VGG-16} model, and a \textit{one-layer LSTM} model are chosen for image and question feature extraction purposes, respectively. The whole network is trained end-to-end by minimizing the categorical loss between the predicted and ground-truth answers.

\par From Table \ref{tab:accu}, we can see that prediction directly from question \textit{Q-Only} is very poor compared to the other VQA approaches. This demonstrates that there is less language bias (i.e., the ability to predict an answer directly from a question) present in our dataset. The other four VQA models perform nearly the same on our \textit{Polar-VQA} dataset. We can identify that overall accuracy lies somewhere between $0.96-0.97$ which is nearly perfect on test data. The accuracy for each category is also close to $1$. These promising results highlight that the VQA framework is highly capable of identifying the types of radars, recognizing locations, and differentiating between dry and wet zone images very efficiently.

\par Table \ref{tab:accu_2} highlights the performance of two types of questions for each question category. We can see that for open-ended (OE) questions, the accuracy is 1 in the case of identifying the dry and wet zone images, regardless of the VQA models. However, the accuracy of the place category is lower than that of the other two categories. On the other hand, for close-ended (CE) questions, performance is better for the zone category among all the question categories. From both Table \ref{tab:accu} and \ref{tab:accu_2}, we understand that performance among the VQA approaches given a question category is less deviated than the performance between question categories.


\section{Conclusion}
\label{conclusion}
\par The visual question-answering task on ice sheet imagery collected using various airborne sensors is introduced in this study. To do so, we have presented a unique VQA dataset, namely \textit{Polar-VQA}. From the comparative study among several VQA approaches, we can see that VQA has the potential to be incorporated into ice sheet research, where an expert can obtain vital information by asking questions. Developing a large-scale dataset that includes more types of questions targeted at extracting several types of information besides radars, locations, and zones is our further research direction. 

\subsubsection*{Acknowledgments}
This work was supported by NSF BIGDATA Awards (IIS-1838230, IIS-1947584), NSF HDR
Institute Award (OAC-2118285), IBM, and Amazon. We also acknowledge the support of the U.S. Army Grant No. W911NF2120076.

\bibliographystyle{IEEEbib}
\bibliography{main}

\end{document}